\documentclass[conference]{IEEEtran}
\usepackage[sc]{mathpazo}
\usepackage[T1]{fontenc}

\usepackage{flushend}
\usepackage[numbers,sort&compress]{natbib}
\usepackage[cmex10]{amsmath}
\usepackage{array} 
\newcolumntype{M}{>{$\vcenter\bgroup\hbox\bgroup}c<{\egroup\egroup$}}
\usepackage{url}
\ifCLASSINFOpdf
  \usepackage[pdftex]{graphicx}
  \DeclareGraphicsExtensions{.pdf,.jpeg,.jpg,.png}
\else
  \usepackage[dvips]{graphicx}
  \DeclareGraphicsExtensions{.eps}
\fi

\usepackage{amssymb}
\usepackage{booktabs}
\usepackage{graphicx}
\usepackage[normalem]{ulem}

\newcommand{\sref}[1]{Sec. \ref{#1}}
\newcommand{\figref}[1]{Fig.\ref{#1}}

\usepackage[font=footnotesize,labelfont=bf]{caption}

\usepackage{ifthen}
\usepackage[usenames,dvipsnames]{color}
\newboolean{include-notes}
\setboolean{include-notes}{true}

\newcommand{\adnote}[1]{\ifthenelse{\boolean{include-notes}}%
 {\textcolor{blue}{\textbf{AD: #1}}}{}}

\DeclareMathOperator*{\argmax}{argmax}

\DeclareFontFamily{U}{MnSymbolC}{}
\DeclareSymbolFont{MnSyC}{U}{MnSymbolC}{m}{n}
\DeclareFontShape{U}{MnSymbolC}{m}{n}{
    <-6>  MnSymbolC5
   <6-7>  MnSymbolC6
   <7-8>  MnSymbolC7
   <8-9>  MnSymbolC8
   <9-10> MnSymbolC9
  <10-12> MnSymbolC10
  <12->   MnSymbolC12%
}{}
\DeclareMathSymbol{\powerset}{\mathord}{MnSyC}{180}

\newcommand{\R}{{\mathcal{R}}}
\newcommand{\Hu}{{\mathcal{H}}}

\newcommand{\uH}{{\bf u}_{\mathcal{H}}}

\newcommand{\uR}{{\bf u}_{\mathcal{R}}}

\newcommand{\rR}{r_\mathcal{R}}

\newcommand{\rH}{r_\mathcal{H}}

\newcommand{\RH}{R_\mathcal{H}}
\newcommand{\RR}{R_\mathcal{R}}

\newcommand{\subh}{_{\mathcal{H}}}

\newcommand{\prg}[1]{\vspace{.5em}\noindent\textbf{#1. }}

\title{Robot Planning with Mathematical \\Models of Human State and Action}
 
\author{{\large  Anca D. Dragan (anca@berkeley.edu)} \\
  {\small Department of Electrical Engineering  and Computer Sciences}\\
 {\small University of California, Berkeley\vspace{.2em}}\\ {\small Summary of work in collaboration with P. Abbeel, R. Bajcsy,} \\{\small  A. Bestick, J. Fisac, T. Griffiths, JK. Hedrick, J. Heimrick, N. Landolfi, C. Liu,} \\{\small D. Hadfield Menell,  S. Milli, A. Nagabaudi, S. Russell, D. Sadigh, S. Sastry, S. Seshia, S. Srinivasa, A. Zhou}}

\begin{document}

\maketitle

\begin{abstract}
Robots interacting with the physical world plan with models of physics.  We advocate that robots interacting with people need to plan with models of cognition. This writeup summarizes the insights we have gained in integrating computational cognitive models of people into robotics planning and control. It starts from a general game-theoretic formulation of interaction, and analyzes how different approximations result in different useful coordination behaviors for the robot during its interaction with people.
\end{abstract}

\section{Introduction}
Robots act to maximize their utility. They reason about how their actions affect the state of the world, and try to find the actions which, in expectation, will accumulate as much reward as possible. We want robots to do this well so that they can be useful to us -- so that they can come in support of real people. But supporting people means having to work with and around them. We, the people, are going to have to share the road with autonomous cars, share our kitchens with personal robots, share our control authority with prosthetic and assistive arms. 

Sharing is not easy for the robots of today. They know how to deal with obstacles, but people are more than that. We reason about the robot, we make decisions, we act. This means that the robot needs to make predictions about what we will think, want, and do, so that it can figure out actions that coordinate well with ours and that are helpful to us. Much like robots of today have a theory of \emph{physics} (be it explicitly as an equation or implicitly as a learned model), the robots of tomorrow will  need to start having a theory of \emph{mind}. 

Our work for the past few years has focused on integrating mathematical theories of mind, particularly about human \emph{future actions} and \emph{beliefs}, into the way robots plan their \emph{physical}, task-oriented actions.

This required a change from the robotics problem formulation (\figref{fig:front}, left), to an \emph{interaction} problem formulation (\figref{fig:front}, right). Interaction means there is not a single agent anymore: the robot and human are both agents in a two player game, and they take actions according to utility functions that are not necessarily identical or known to each other. The paper outlines this formally in \sref{sec:game}, and then summarizes the different approximations we've explored and what we've learned from them \cite{Sadigh_RSS_driving,Chang_AAMAS_goalinference,Bestick_ISER_influence,Sadigh_comparison,Sadigh_IROS_infogather,hadfield2016cooperative,hadfield2017ird,Milli_obedience,Menell_offswitch,Fisac_WAFR_tpred,Dragan_RSS_legibility,Zhang_HRI,Huang_expressrewards}:

\begin{figure}
\includegraphics[width=\columnwidth]{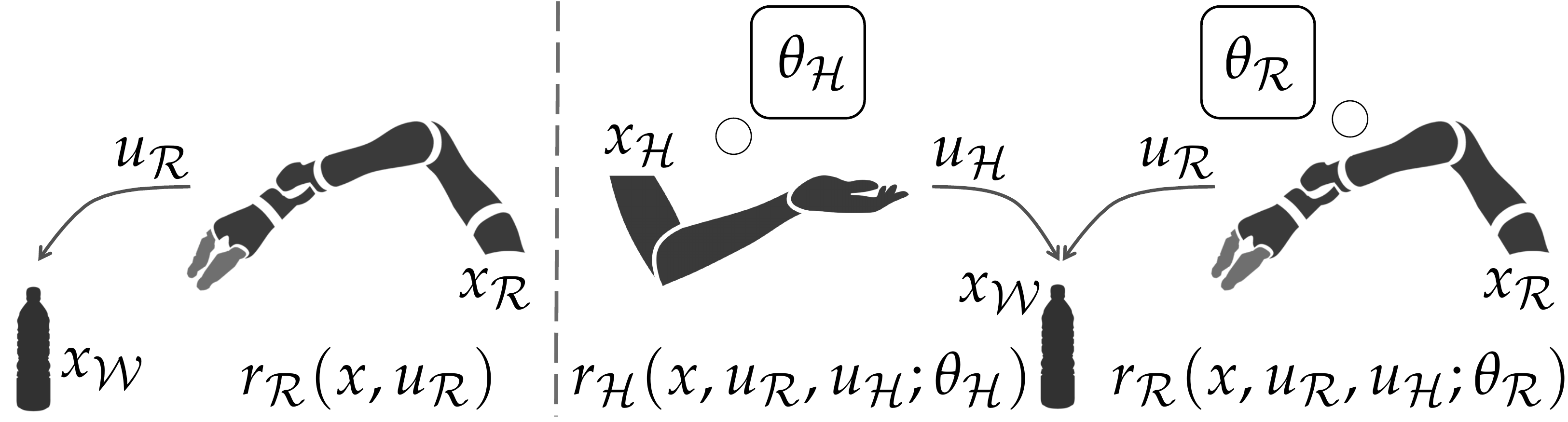}
\caption{Left: Traditional robotics formalism: the robot takes actions $u_{\mathcal{R}}$ to optimize a reward or cost function $r_{\R}$. Right: Interaction formalism: the robot is not acting in isolation; the human takes actions $u_{\Hu}$ to optimize a reward function $r_{\Hu}$, possibly different from that of the robot's, and parametrized by human internal/hidden state $\theta_{\Hu}$; the human does not know the robot internal state $\theta_{R}$, which parametrizes the robot's reward function; and both functions depend on both agents' actions.}
\label{fig:front}
\end{figure}

\prg{Accounting for the \emph{physical human behavior} during interaction (\sref{sec:humanbehavior})} \begin{quote}\emph{One important insight for us has been that people can't be treated as just obstacles that move: they will \textbf{react} to what the robot does.}\end{quote} If a car starts merging in front of you, you break. If the robot helping you assemble a part employs a different strategy than you expected, you adapt. It took more and more sophisticated approximations to the game above to account for this. 

Our first approximation to the game started by assuming a shared utility function and treating the person as a perfect collaborator, but replanning at every step to adapt to when the person deviates from the collaborative plan \cite{Chang_AAMAS_goalinference}; we then relaxed this to an imperfect collaborator model, showing that the robot can leverage its actions to guide the person to perform better in the task \cite{Bestick_ISER_influence}; finally, we investigated a model of the person as optimizing a different utility function, but simply computing a best response to the robot's actions (as opposed to solving the full dynamic game) \cite{Sadigh_RSS_driving} -- this model enables the robot to account for how people will react to its actions, and thus perform better at its task. 

\prg{Using the human behavior to infer \emph{human internal states} (\sref{sec:inferhuman})} The models above were a first step in coordinating with people, but they were disappointing in that they still assumed perfect information, i.e. everything is known to both parties. It is simply not true that we will be able to give our robots up front a perfect model of each person they will interact with. Next, we studied how robots might be able to estimate internal, hidden, human states, online, by taking human behavior into account as evidence about them.

\begin{quote}
\emph{Another important insight has been that robots should not take their objective functions for granted: they are easy to misspecify and change from person to person. Instead, robots should optimize for what the person wants internally, and use human guidance to estimate what that is.}
\end{quote}

Inverse Reinforcement Learning \cite{ng2000algorithms,ziebart2008maximum,ramachandran2007bayesian} already addresses this for a \emph{passive} observer analyzing \emph{offline} human \emph{demonstrations} of approximately optimal behavior \emph{in isolation}. Our work builds on three goals: 1) making these inferences \emph{actively} and \emph{online}; we leverage not just queries \cite{Sadigh_comparison}, but also the robot's physical actions \cite{Sadigh_IROS_infogather}; 2) accounting for the fact that if the person knows the robot is trying to learn, they will act differently from what they do in isolation and \emph{teach} \cite{hadfield2016cooperative}; and, perhaps most importantly, 3) leveraging richer sources of human behavior beyond demonstrations, like physical corrections \cite{bajcsy2017phri}, orders given to the robot \cite{Milli_obedience}, and even the reward function specified \cite{hadfield2017ird} -- all of these are observations about the actual desired reward.

We argue that interpreting the reward function designed for the robot as useful information about the true desired reward, by leveraging the context in which this function was designed to begin with, can make robots less sensitive to negative consequences of misspecified rewards \cite{hadfield2017ird}.  Similarly, we find that interpreting orders as useful information about the desired reward give robots a provable incentive to accept human oversight as opposed to bypass it in pursuit of some misspecified objective \cite{Menell_offswitch}. 
Overall, we find that accepting that the robot's reward function is not given, but part of the person's internal state, is key to safe and customizable robots. 

\prg{Accounting for \emph{human beliefs} about robot internal states (\sref{sec:beliefs})} 
\begin{quote}\emph{Robots need to make inferences about people during interaction, but people, too, need to make inferences about robots. Robot actions influence whether people can make the correct inferences.}
\end{quote}
The third part of our work focuses on getting robots to produce behavior that enables these human inferences to happen correctly, whether they are about the robot's behavior \cite{Fisac_WAFR_tpred}, or about the robot's internal states (like utility \cite{Huang_expressrewards}, goals \cite{Dragan_RSS_legibility}, or even level of uncertainty \cite{Zhang_HRI}). Although these increases the robot's transparency, we have been encoding the need for that in the objective directly, whereas really it should be a consequence of solving the interaction problem well. This is something we are actively looking into, but which increases the computational burden.

\vspace{.5em}

All this research stands on the shoulders of inspiring works in computational human-robot interaction, nicely summarized in sections 6 and 7 of \cite{thomaz2016computational} and not repeated here. What this writeup contributes a summary of our own experiences in this area, particularly focusing on physical, task-oriented interaction. It provides a common formalism that, in retrospect, can be seen as the general formulation that seeded these works, along with a quasi-systematic analysis of the different ways to approximate solutions and the sometimes surprisingly interesting and powerful behavior that emerges when we do that. This enables robots to \emph{tractably and autonomously generate} different kinds of behavior for interaction, often \emph{in spite of the continuous state and action spaces} they have to handle -- from arms that guide and improve human performance in handovers, to cars that negotiate at intersections, to robots that purposefully make their inner-workings more transparent to their end-users.

\begin{figure*}
\includegraphics[width=\textwidth]{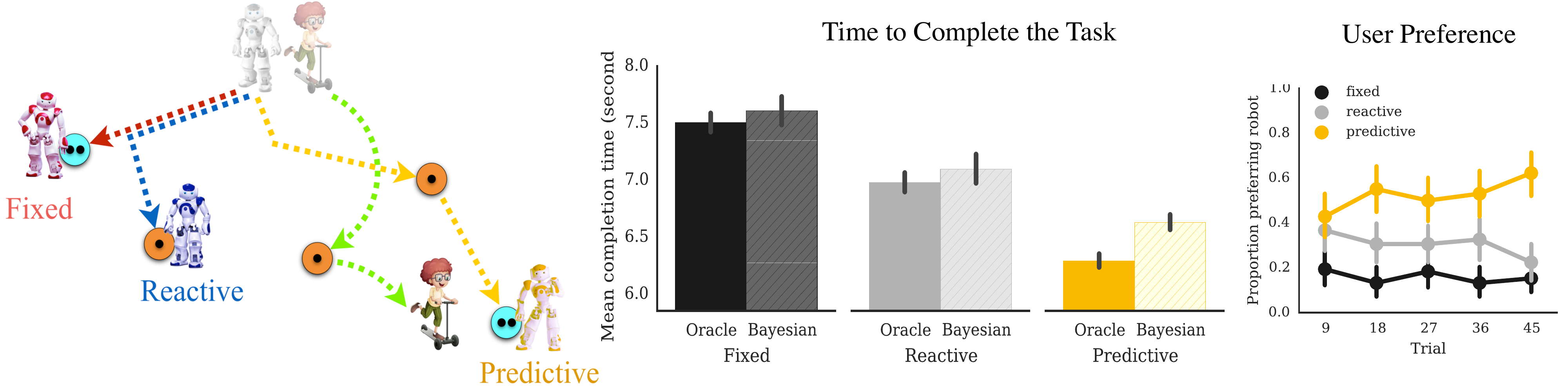}
\caption{The Human as a Collaborator: On online study on collaborative Traveling Salesman Problems. The robot models the person's behavior as rational w.r.t. to the same reward function as its own reward function. In the "Fixed" condition, the robot plans an optimal centralized plan for both the human and the robot, and executes its portion. If the human has a different plan, they end up needing to adapt. In the "Reactive" condition, the robot adapts its plan after every target the human reaches, recomputing the new optimal centralized plan from that new state. This performs significantly better than the human plan in objective and subjective task performance measures. Finally, in the "Predictive" condition, the robot uses Bayesian Inference to predict the human's next target via an observation model based on the rationality assumption, and can proactively replan before the human reaches their target.  }
\label{fig:collaboration}
\end{figure*}

\section{General Interaction as a Game}\label{sec:game}
In general, we can formulate interaction as a 2-player dynamic game between the human and the robot. The state $x$ contains the world state along with the robot and human state:
$$x=(x_{\mathcal{W}},x_{\R},x_{\Hu})$$
Each agent can take actions, and each agent has a (potentially different) reward function: $$\rR(x,u_{\R},u_{\Hu};\theta_{\R})$$ for the robot and  $$\rH(x,u_{\R},u_{\Hu};\theta_{\Hu})$$ for the human, each with parameters $\theta$. The two do not necessarily know each other's reward functions (or equivalently, each other's parameters $\theta$).

Let $T$ be the time horizon, and $\uR$ and $\uH$ be a robot and human, respectively, control sequence of length $T$. We denote by $\RR$ the cumulative reward to the robot from the starting state $x^0$:
$$ \RR(x^0,\uR,\uH;\theta_{\R})=\sum_{t=0}^T \rR(x^t,u_{\R}^t,u_{\Hu}^t;\theta_{\R})$$
and
by $\RH$ the cumulative reward for the human:
$$ \RH(x^0,\uR,\uH;\theta_{\Hu})=\sum_{t=0}^T \rH(x^t,u_{\R}^t,u_{\Hu}^t;\theta_{\Hu})$$

One way to model what the person will do is to model them as rationally solving this game. There are several issues with this. The first is that it is computationally intractable. The second is that if $\rH\neq\rR$, the game will have many equilibria, so even if we could compute all of them we'd still not be sure which the person is using.  The third is that even without the first two issues, this would still not be a good model for how  people make decisions in day to day tasks \cite{hedden2002you,rubinstein1998modeling}. 

Our work has thus explored different approximations to this problem that might better match what people do, while enabling robots to actually generate their actions in realistic tasks in (close to) real time.

\section{Human Behavior During Interaction}\label{sec:humanbehavior}

Because in interaction the robot's reward depends on what the person does, the ability to anticipate human actions becomes crucial in deciding what the robot should do. Rather than modeling the person as solving the game above, we explored several approximations that each led to different yet useful behaviors in interaction.

\prg{The Perfect Collaboration Model}
The easiest simplifying assumption is actually that the person is optimizing the same reward function:
$$\rH=\rR$$
and both agents know this (no more partial information). 
This assumption turns planning for interaction into a much easier problem, analogous to the original robotics problem: it is pretending like the person is just some additional degrees of freedom that the robot can actuate -- their actions will follow the optimal centralized plan:
$$(\uR^{*},\uH^{*})=\argmax_{\uR,\uH}\RR(x^{0},\uR,\uH)$$

Despite its simplicity, we have found that this can be very useful so long as the robot \emph{replans} at every time step. People inevitably deviate from from the optimal centralized plan even from the first step, ending up in some new state -- because they don't actually optimize the same reward, because they don't know that the robot is optimizing the same reward, or because they are not perfect optimizers. But the robot can recompute the centralized optimal plan from that new state, and proceed with the first action from the new $\uH^{*}$. 

\figref{fig:collaboration} shows a comparison from \cite{Chang_AAMAS_goalinference} between a ``Fixed'' robot strategy, where the robot executes the originally planned $\uR^{*}$ regardless of what the person does, and a ``Reactive'' strategy, in which the robot keeps updating $\uR$ based on the new centralized optimal plan from the current state at every step. Here, $\RH=\RR$ and equates to the total time to solve the task. We recruited 234 participants on Amazon Mechanical Turk who played collaborative Traveling Salesman Problems with a robot avatar in a within-subjects design that included a 3rd condition, discussed later, and a randomized trial order. We found that the Reactive condition led to the task being completed significantly faster by the human-robot team, and that participants preferred the Reactive robot especially in the beginning. 

\begin{figure*}
\includegraphics[width=\textwidth]{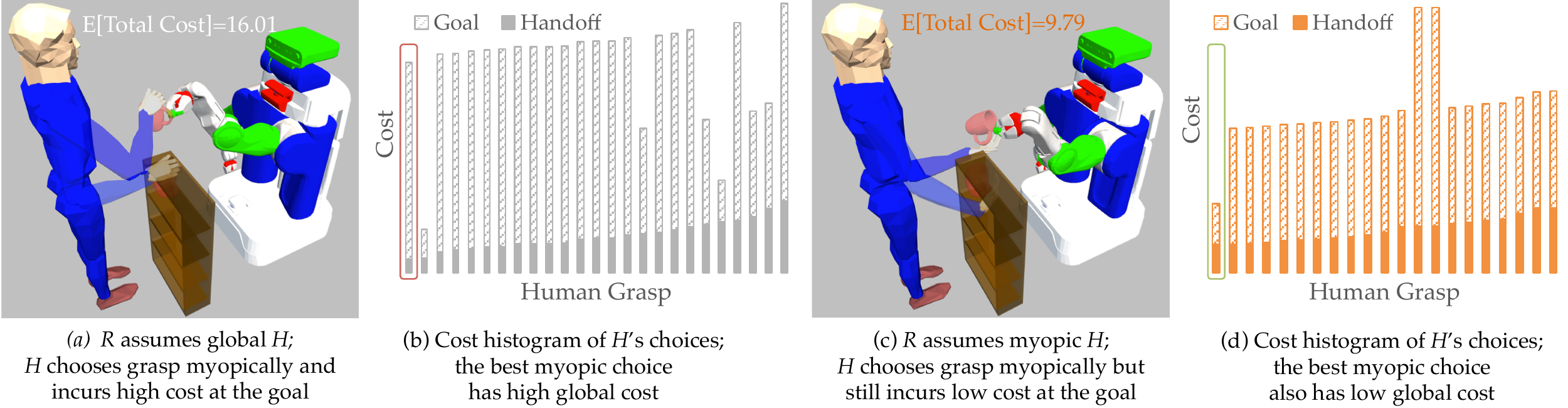}
\caption{People can be myopic about their decision, greedily optimizing reward as opposed to looking ahead (a), e.g. choosing the most comfortable way to grab a mug even though it they would have to regrasp it in order to set it down in a shelf. This might mean that the best action locally has poor reward globally (b). When accounting for human myopia, the robot can select actions such that the myopic response is still good globally (c,d).}
\label{fig:handover}
\end{figure*}
\emph{Overall, online planning with a perfect collaborative model of human behavior can be useful, already enabling the robot to continually adapt to the person's plan even though it does not get it right a priori. }

\prg{Collaborative but Approximately Optimal}
An improvement upon the perfectly rational collaborator human model is to recognize that people are not actually perfectly rational. In \cite{Bestick_ISER_influence}, we model people as collaborative still, but no longer assume $\uH^{*}$ is perfect. In particular, we assumed that people are \emph{greedy} or \emph{myopic} in their decisions. Their action at time $t$ will be the one that looks locally good, not the one that is optimal over the full time horizon:
$$u_{\mathcal{H}}^{t*}(u^{t}_{\mathcal{R}})=\arg\max_{u^{t}_{\mathcal{H}}}\rR(x^t,u^{t}_{\mathcal{R}},u^{t}_{\mathcal{H}})$$

The robot can then choose its own actions such that, when coupled with the person's myopic response to them, the combined plan is as high-reward as possible:
$$\uR^{*}=\arg\max_{\uR}\RR(x^{0},\uR,\uH^{*}(\uR))$$

This results in the robot \emph{guiding} the person's actions, helping them overcome cognitive, bounded-rationality limitations as much as possible.

Interaction becomes an \emph{underactuated system}: the robot no longer assumes it can directly actuate $\uH$, but accounts for how $\uR$ will influence $\uH$ and takes that into account when planning.

In particular, we investigated a handover task in which participants had to take an object from the robot and place it at a goal location. We used as $\rH=\rR$ as negative ergonomic cost to the person. People's decision in this problem is how to take the object such that it is ergonomically low cost. The robot's decision is how to hold the object at the handover time to enable that.  A perfect human optimizer would minimize cost at the handover \emph{and} at the goal. Our myopic model minimized cost the handover time, which then could in some cases resulted in high cost at the goal, such as needing to twist their arm in an uncomfortable way. or regrasp. \figref{fig:handover} shows an example. 

When the robot optimizes for its actions,  it chooses ways to hold the object that \emph{incentivize} good global human plans. The robot chooses grasps such that even when the person chooses their grasp greedily for the handover, that greedy grasp is also as close as possible to the global optimum, resulting in low cost (high reward) at the goal as well (\figref{fig:handover}).

\emph{Overall, planning with a myopic collaborative model of the human behavior results in the robot taking actions that can guide the person towards plans that are globally optimal, helping them overcome the limitations of greedy action selection. } 

\prg{Non-Collaborative but Computing a Best Response to the Robot} Not every situations is collaborative. Take driving for example. The car has the same objective as its passenger, but a different objective from other human driven vehicles on the road -- these are trying to reach their own destinations as efficiently as possible, and that sometimes competes with the car's objective.

Breaking the collaboration assumption that the human is optimizing the same reward function as the robot (or viceversa), puts making prediction of human behavior back to solving a 2-player game even if we assume known reward parameters. In \cite{Sadigh_RSS_driving}, we introduced a model that alleviates this difficulty by assuming that the person is not computing to a Nash equilibrium, but instead computing a \emph{best response} to the robot's plan using a different, yet known reward function $\rH$. That is, rather than trying to influence the robot's behavior, the person is taking the robot's behavior as fixed, and optimizing their own reward function within that constraint:
$$\uH^{*}(\uR)=\arg\max_{\uH}\RH(x^{0},\uR,\uH)$$
The robot can then compute the action sequence that, when combined with the human's response to that sequence, leads to the highest value for its \emph{own} reward:
$$\uR^{*}=\arg\max_{\uR}\RR(x^{0},\uR,\uH^{*}(\uR))$$
It can then take the first action in $\uR$, observe the change in the world, and replan. This is what we did in the collaborative replanning case with the Reactive robot, except now the robot has a model for how the person will respond to its actions as opposed to computing a joint global plan. 

We applied this to autonomous driving, namely the interaction between an autonomous car and a human-driven vehicle. Both the robot and the person want to achieve their goal efficiently, which made their reward functions non-identical.  We gave the robot access to $\RH$ by learning it offline using IRL.

\begin{figure*}
\includegraphics[width=\textwidth]{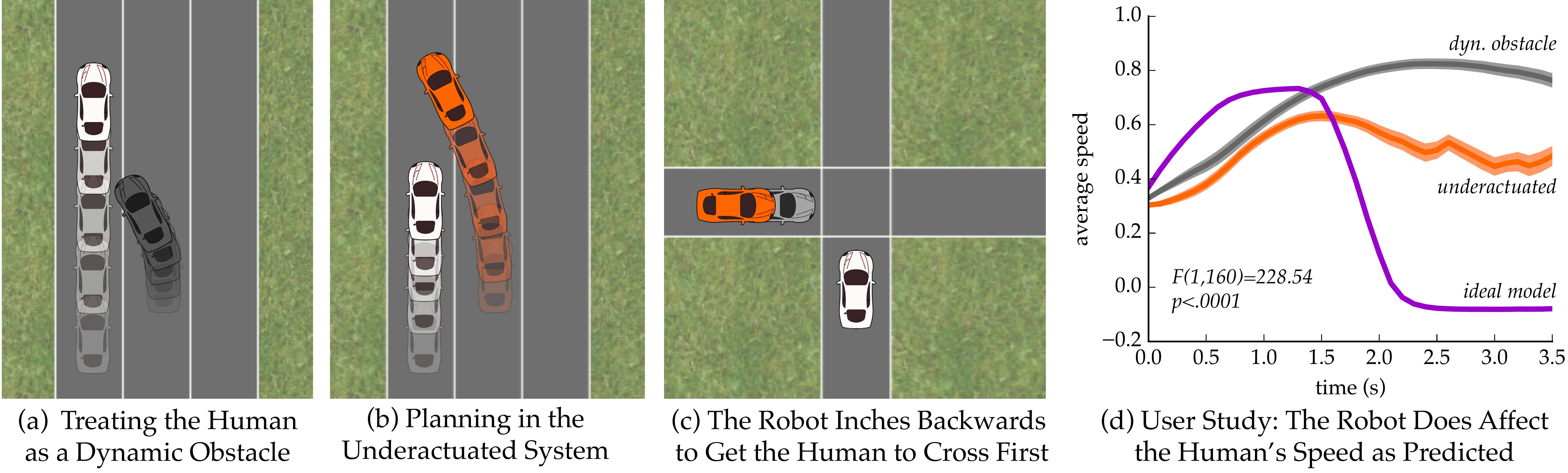}
\caption{Coordination behavior emerges out of not treating people like obstacles that are moving, but modeling how they will rationally react to robot actions.}
\label{fig:efficiency}
\end{figure*}

Typically in autonomous driving, cars treat people like obstacles, planning to stay out of their way. This leads to overly conservative behavior, like cars never getting to merge on a highway, or getting stuck at 4-way stops. In contrast, our car coordinates with people (\figref{fig:efficiency}). It sometimes plans to merge in front of them knowing that they can slow down to accommodate the merge. Or at an intersection, for $\RR$ being higher if the person goes first through the intersection, the robot does not just sit there, but coordinates by \emph{starting to back up}, which makes it safer for the person to go (effectively \emph{signaling} to the human driver). We ran a user study with a driving simulator, and the results suggested that people's behavior when the robot is planning with this model leads to significantly higher reward for the robot than in the baseline of treating the person as on obstacle. 

\emph{Overall, treating interaction as an underactuated system where the person is not playing a game, but acting rationally as a best response to the robot's actions, leads to coordination behavior that naturally emerges out of the optimization over robot actions.}

\section{Using Human Behavior \\to Infer Human Internal State}\label{sec:inferhuman}
Thus far we've oversimplified the game by assuming that the robot knows the persons's reward parameters. In reality, the robot does have direct access to these. Even further, we argue that in reality, the robot does not really have access to its own reward parameters either -- collaborative robots, meant to help a person, should optimize for whatever that person wants, not for some a-priori determined reward function. 

Robots today take their reward function as given. But where does this reward function come from? Typically, it is designed by some person who does their best at writing down what they think the robot should optimize for. Unfortunately, we, people, are terrible at specifying what we want. From King Midas to The Sorcerer's Apprentice, we have countless stories that warn us about unintended, negative consequences of misspecified wishes or objectives. We propose that robots should have uncertainty over their objectives, and that they should try to optimize for what people want internally, but can't necessarily explicate. This is key to alleviating the negative consequences of a misspecified objective.

To achieve this, we use human actions as observations about their internal or desired reward function. 
as they would in the full dynamic game. But the robot will no longer assume that it knows $\rH$. However, if we assume a rational model of human behavior, then the human actions become observations about this hidden internal human state. 
To estimate $\theta_{\Hu}$ from human actions, the robot needs an observation model -- the probability of observed actions given $\theta_{\Hu}$. We assume the person is approximately rational \cite{ziebart2008maximum,baker2007goal} with a model that comes from a maximum entropy distribution in which trajectories are more likely when their total reward is higher:
$$P(\uH|x^{0},\uR,\theta_{\Hu})\propto \exp({\RH}(x^{0},\uR,\uH;\theta_{\Hu}))$$

Then the robot can update its belief over $\theta_{\Hu}$:
$$b'(\theta_{\Hu})\propto b(\theta_{\Hu})P(\uH|x^{0},\uR,\theta_{\Hu})$$ 

If the robot observes a trajectory $\uH$ for the full time horizon, then the belief update equates to (Bayesian) Inverse Reinforcement Learning \cite{ziebart2008maximum,ramachandran2007bayesian}. 

But robots sometimes need to infer the human reward online, as the human trajectory is unfolding. Think back to the driving application: it is not the case that every person optimizes the same reward function: some drivers are more aggressive than others, some are not paying attention, and so on. It is therefore helpful to be able to update $\theta$ as the robot is interacting with the person. 
In such cases, the robot has only observed $\uH^{0:t}$ and must update its belief based just that, rather than a full trajectory. 

Further, robots have an opportunity to go beyond passive inference, and use their actions for \emph{active} estimation, triggering informative human reactions. Instead of passively observing what people do, they can leverage their actions to gather information. 

Finally, it's not just human physical actions that are informative. Human behavior in general, like physical corrections, comparisons, orders given to the robot, and even a reward function that  a designer tries to write down -- all of these are useful sources of information about what the true robot objective should be.

\prg{Online Inference by Integrating over Futures}
In \cite{Chang_AAMAS_goalinference,Dragan_RSS_teleop}, we integrated over the possible future human trajectories in order to compute the belief update:
$$ P(\uH^{0:t}|x^{0},\uR,\theta_{\Hu})=\int P(\uH|x^{0},\uR,\theta_{\Hu}) d\uH^{t+1:T}$$
We showed that for the case of the reward $\rH$ being parametrized by which goal $\theta_{\Hu}$ the person is reaching for, the integral can be approximated via Laplace's method, assuming reward 0 for trajectories that do not reach the goal.

\figref{fig:collaboration} shows a Predictive condition from \cite{Chang_AAMAS_goalinference}, in which the robot infers the person's goal and uses it to proactively change its plan. This condition outperforms the Reactive condition. In \cite{Dragan_RSS_teleop}, we used goal inference to adapt to an operator's goal during teleoperation of a robot arm.

\emph{Overall, human actions as observations about the underlying human goals, and inferring these enables robots to proactively adapt to what people want.}

\prg{Active Online Inference using Robot Physical Actions}
Inference does not have to be passive. In \cite{Sadigh_comparison}, we explored active inference, where the robot makes queries that the person responds to. But really, having a robot whose actions influence human behavior presents an opportunity: to leverage robot actions and trigger informative human reactions. 

In \cite{Sadigh_IROS_infogather}, we ran online inference by modeling the human as optimizing with a shorter time horizon. We then took the human (short time horizon) trajectory as evidence about the underlying $\theta_{\Hu}$. 
There, $\theta_{\Hu}$ parametrized the reward function by representing weights on different important features of the human state and action. Unlike goals, this is a continuous and high-dimensional space.
So rather than maintaining a belief over all possible $\theta_{\Hu}$s, we clustered users into styles and only maintained a belief over a discrete set of driving styles.  

Further, we sped up the inference by leveraging the robot's actions: since the person will choose actions that depend on what the robot does, $\uR$, the robot has an opportunity to select actions that \emph{maximize information gain} (trading off with maximizing reward using the current estimate $\hat{\theta}_{\Hu}$:
\begin{align}
\uR^{*}=\arg\max_{\uR}\RR(x^{0},\uR,\uH^{*}(\uR;\hat{\theta_{\Hu}}))\nonumber \\ 
+\lambda (H(b)-\mathbb{E}_{\theta_{\Hu}}H(b')) \nonumber
\end{align}


Note that if we were able to treat $\theta_{\Hu}$ as the hidden state in a POMDP with very complicated dynamics (that require planning for the person to solve for how the state will update given the robot's action), then the robot's policy would achieve an optimal trade-off between exploiting current information and gathering information. However, since even POMDPs with less complex dynamics are still intractable in continuous state and action, we resorted to an explicit trade-off. 

We found that the robot planning in this formulation exhibited interesting behavior that could be seen as information-gathering. For instance, it would nudge closer to someone's lane, because the anticipated reaction from the person would be different depending on their driving style: attentive drivers break, distracted drivers continue. Or, at a 4-way stop, it would inch forward into the intersection, again anticipating different reactions for different styles. 

\emph{The robot can leverage its actions' influence on the person to actively gather information about their internal reward parameters.}

\prg{What If the Human Knows the Robot is Learning?}
One issue with estimation arises when the observation model is wrong. People might act approximately optimal with respect to the reward function, except when they know that the robot is trying to learn something from them. This is why coaches are different from experts: when we teach, we simplify, we exaggerate, we showcase. A gymnastics coach does not demonstrate the same action they would perform if they were in the olympics. 

In \cite{hadfield2016cooperative}, we analyzed the difference between maximizing the reward function for the true $\theta_{\Hu}^*$:
$$\uH^{expert}=\arg\max_{\uH} \RH(x^0,\uR,\uH;\theta_{\Hu}^*)$$
and maximizing the probability that the robot will infer the true $\theta_{\Hu}^*$:
$$\uH^{teacher}=\arg\max_{\uH}b'(\theta_{\Hu}^*)=\arg\max_{\uH}P(\theta_{\Hu}^*|x^0,\uR,\uH) $$

\begin{figure}[h!]
\includegraphics[width=\columnwidth]{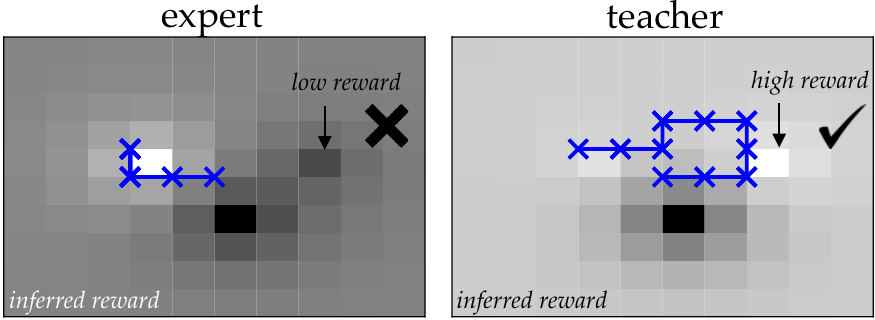}
\caption{We model teaching demonstrations as being informative about the underlying reward function. An expert demonstration (left) might lead to inferring the wrong reward function, e.g. because the expert goes straight for the high reward peak nearby. A teaching demonstration will deviate from optimality to showcase the underlying reward function, e.g. the teacher goes to both high reward peaks to clarify.}
\label{fig:cirl}
\end{figure}

\figref{fig:cirl} compares $\uH^{expert}$ to $\uH^{teacher}$ in a simple MDP where the reward function consists of high and low reward peaks. The expert demonstration heads straight to the closest high reward peak, but the robot has trouble inferring that there is another peak with also high reward. In contrast, the teaching demonstration visits both, leading to the robot inferring the correct $\theta_{\Hu}$.

\emph{Overall, we should expect and account for how people will act differently when they are trying to teach the robot about their internal reward parameters.}

\begin{figure}[h!]
\includegraphics[width=\columnwidth]{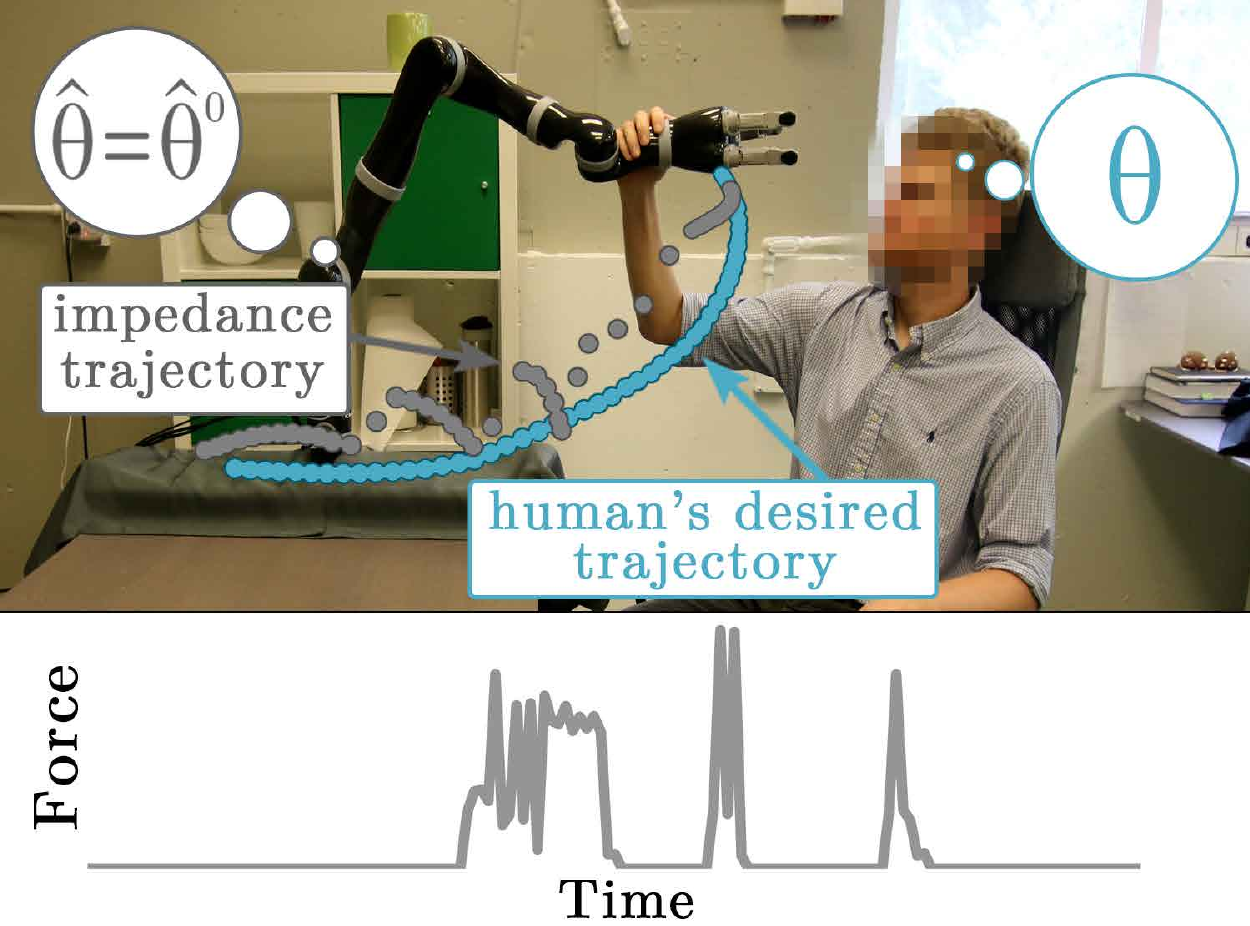}
\includegraphics[width=\columnwidth]{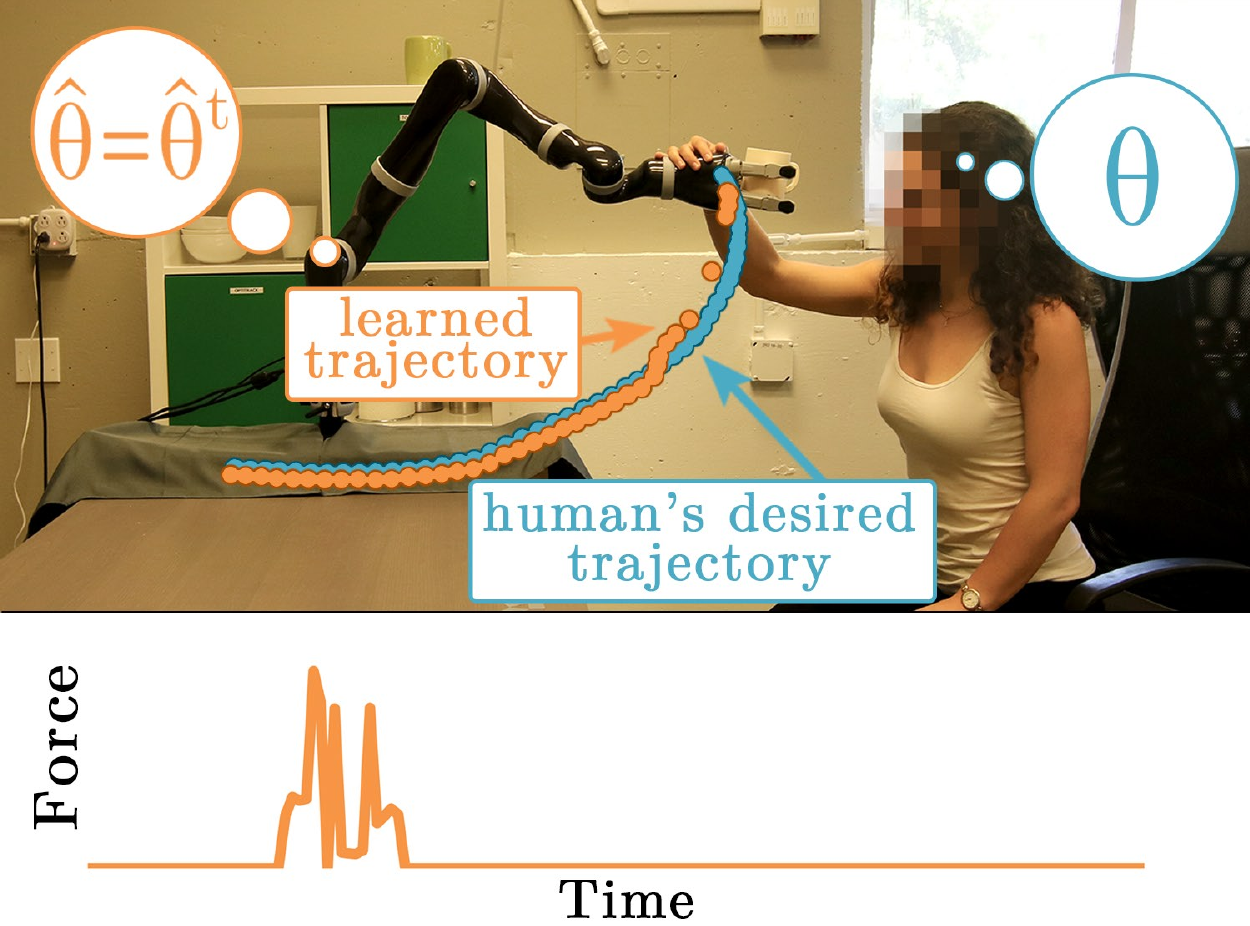}
\caption{Learning the desired objective function from physical corrections, in real-time, leads to completing the task in the desired way with less human intervention.}
\label{fig:phri}
\end{figure}

\prg{Learning Objectives from Rich Human Guidance} It is not just human physical actions as part of the task that should inform the robot about the internal human objective. We explored physical corrections \cite{bajcsy2017phri} (\figref{fig:phri}), comparisons \cite{Sadigh_comparison}, orders (human oversight) \cite{Milli_obedience,Menell_offswitch}, and even attempts at specifying an objective \cite{hadfield2017ird}, all as sources of information for the robot. Each required its own observation model, and its own approximations for running the inference. 

In \cite{hadfield2017ird}, we proposed to mode the reward design process: the probability that a reward designer would choose $\theta_{\R}$ as the specified reward, given the true reward $\theta^{*}$ \emph{and} the training environments they are considering. We then showed how the robot can invert this model to get a posterior distribution over what the true reward is, and that this alleviates consequences like reward hacking and negative side-effects. Surprisingly, we found that this works even when the important features affecting the true reward, like the presence of a dangerous kind of terrain, are latent and not directly observed. 

Even more surprising is our finding from \cite{Menell_offswitch}, where we focused on shut down orders to the robot. Intuitively, robots should just follow such orders as opposed to try to infer the underlying reward function that triggered them. Unfortunately though, designing a reward function that incentivizes accepting orders is challenging, and so is writing down the right hard constraints that the robot should follow. Instead, our work has proved that when the robot treats orders as a useful source of information about its objective, the incentive to accept them is positive.

\emph{Overall, we find that treating human guidance and oversight as a useful source of information about the robot's true reward can alleviate the unintended consequences of misspecified robot reward functions. }

\section{Human Inferences about the Robot}\label{sec:beliefs}
The previous sections made approximations in which the person knew everything they needed about the robot -- they were computing a best response to the robot and got access to the robot's planned trajectory $\uR$. Here, we relax this. Much like robots do not know everything about people and make inferences about their reward function or goals $\theta_{\Hu}$, people too will not know everything about robots and will try to make similar inferences when deciding on these actions.

Humans interacting with robots will have some belief about $\theta_{\R}$. This section focuses on how robot actions affect not just human actions, but also these human beliefs. This means the robot can specifically choose actions to guide these beliefs towards being as accurate as possible, so that the human actions that follow are well informed. These robot actions end up \emph{communicating} to the human, expressing the robot's internal state.

As of now, we modify the robot's objective to explicitly incentivize communication. We are actively working on making this communication emerge out of the robot optimizing for its own reward function, but now with this more sophisticated model of the person -- one in which robot actions affect human beliefs, and human beliefs are what affect human actions.

\subsection{Humans Expecting Robot Behavior\\ to Be Approximately Rational}
A simple but important inference that people make when observing other agents is what they expect the agent's actions to be. The principle of rational action \cite{sodian2004infants} suggests that people expect rational agents, such as robots, to be rational -- to maximize their own reward:
$$P_{\mathcal{H}}(\hat{\uR}|x^{0})\propto e^{\hat{\RR}(x^{0},\uR)}$$
Here, $\hat{\uR}$ is the person's estimate of what the robot's action sequence will be, and $\hat{\RR}$ is the person's estimate of what the robot's reward function is.

\begin{figure}
\includegraphics[width=\columnwidth]{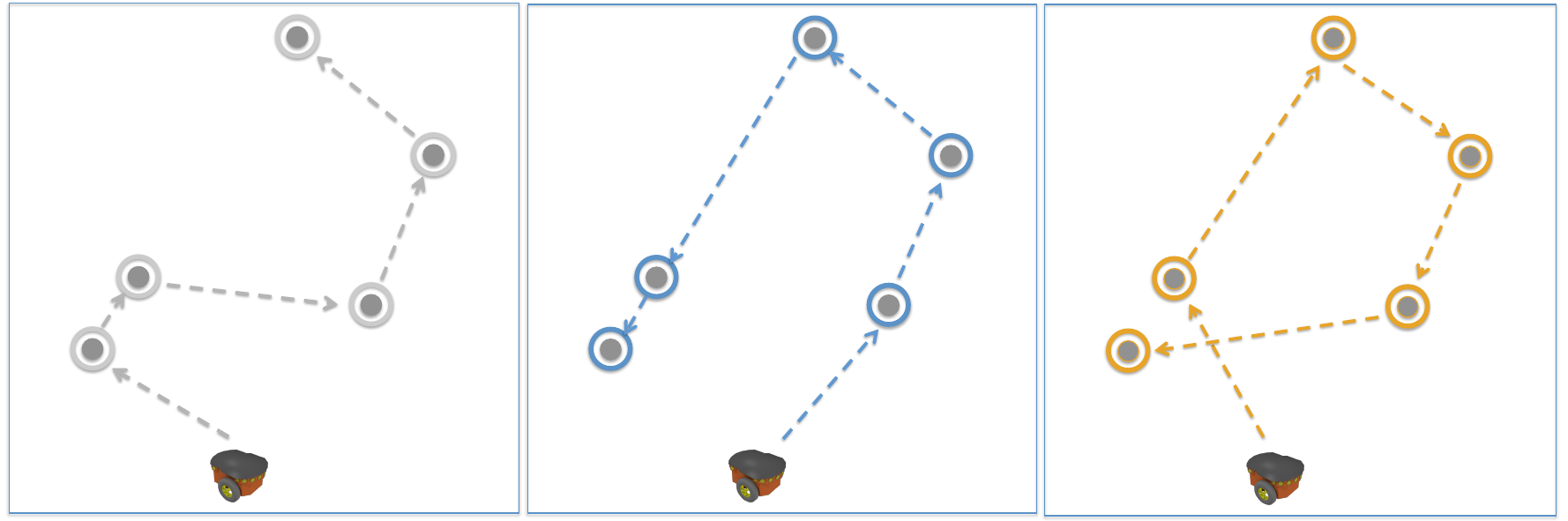}
\caption{After seeing the first $t$ actions, the person should be able to infer confidently the remaining ones. Imagine seeing the first step of the most efficient plan, on the left. It is clear what the robot's 2nd action will be, but after that there are two courses that are relatively close in efficiency. On the other hand, with the middle plan, the first action only leaves one remaining plan sensible. It sacrifices efficiency to make the final T-1 actions clear.  }
\label{fig:wafr}
\end{figure}

Our work leveraged this to generate plans that match what people expect. Namely, at a time $t$, after the person has observed $\uR^{0:t}$, we can model what they expect the robot to do next as
$$P_{\mathcal{H}}(\hat{\uR}^{t+1:T}|x^{0},\uR^{0:t})\propto \exp({\hat{\RR}(x^{0},\uR^{0:t})}+\hat{\RR}(x^{t+1},\uR^{t+1:T}))$$

The robot can use this model to choose a full time horizon plan or trajectory such that the beginning of the plan is informative of the remaining plan, i.e. makes the remaining plan have high probability:
$$\text{\emph{t}-predictability}(\uR)=P_{\mathcal{H}}({\uR}^{t+1:T}|x^{0},\uR^{0:t})$$
$$\uR^{*}=\arg\max_{\uR} \text{\emph{t}-predictability}(\uR)$$
Note that the robot, when interested in its plan being $t$-predictable, might purpusefully deviate from the optimum with respect to $\hat{\RR}$ in order to make sure that the remainder of the plan is what the person would predict after observing $t$ time steps.

\figref{fig:wafr} shows plans optimized for different $t$: 0, 1, and 2. The $t=0$ one is the most efficient. The problem is that after seeing the first step, there are two possible plans that are relatively efficient, so this plan does not do a great job collapsing the person's belief over what will happen, even after they have seen some of the trajectory. In contrast, for $t=1$, this is no longer the most efficient, but makes it very clear what the remainder of the plan will be. \cite{Fisac_WAFR_tpred} details our user studies, both online and in person, with results suggesting that people have an easier time coordinating with robots that are more \emph{t}-predictable.

\emph{Overall, the robot can leverage the person's expectations about its actions to make its plans more predictable.}

\subsection{Humans Using Robot Behavior \\to Infer Robot Internal State}

Once people have a model of how the robot will behave, they can also start using that model to perform inference about hidden states, like the robot's goals or objectives. 

Building on \cite{baker2007goal,csibra2007obsessed}, we have been exploring Bayesian Inference as a model of how people infer robot internal state $\theta_{\R}$ from observed robot actions. This model is analogous to the algorithms we used in the human behavior section to enable robots to infer human internal state from observer human actions:
$$P_{\mathcal{H}}(\uR|x^{0},\theta_{\R})\propto \exp({\hat{\RR}}(x^{0},\uR;\theta_{\R}))$$
$$b\subh'(\theta_{\R})\propto b\subh(\theta_{\R})P_{\mathcal{H}}(\uR|x^{0},\theta_{\R})$$ 

The robot can now communicate a $\theta_{\R}^{*}$:
$$\uR^{*}=\arg\max_{\uR} b\subh'(\theta_{\R}^{*})$$
This is analogous to pragmatics, but the communication happens through physical behavior and not through language.  

\begin{figure}[h]
\includegraphics[width=\columnwidth]{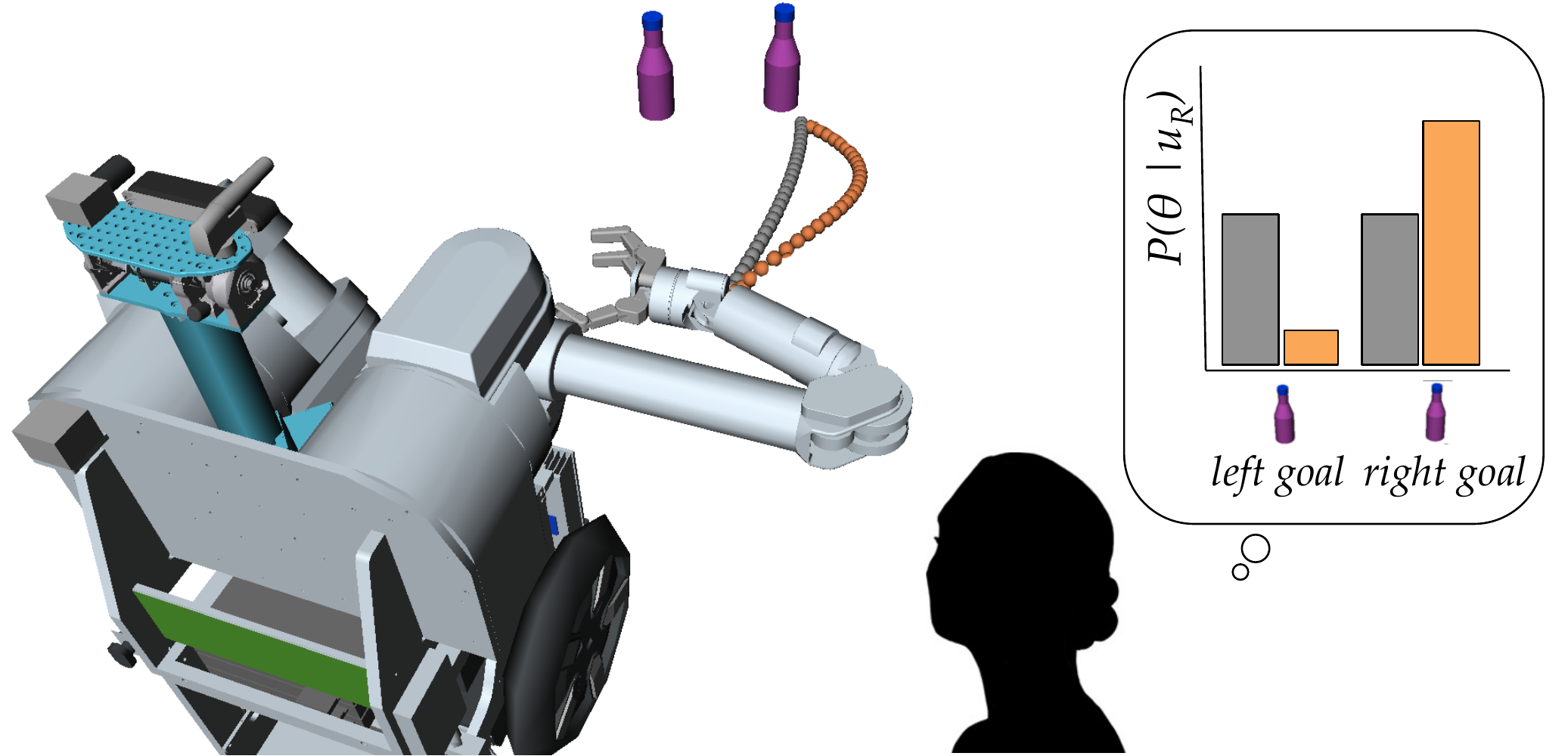}
\caption{The robot models the person as running Bayesian Inference to infer its goal. It chooses to exaggerate its motion to the right to convey that its goal is the bottle on the right. }
\label{fig:legibility}
\end{figure}

\prg{Communicating Robot Goals}
In earlier work \cite{Dragan_RSS_legibility}, we studied a version of this formulation where $\theta_{\R}$ is the robot's goal. A manipulator arm decides to exaggerate its trajectory to the right to convey that the correct goal is the one on the right and not the one on the left (\figref{fig:legibility}), and this does lead to participants' inferring the robot's goal faster. 

The human inference model when $\theta_{\R}$ is a goal is one that has been heavily explored in cognitive science (e.g. \cite{baker2007goal}), and this work showed what happens when a robot uses it for communication. The result is analogous to findings in human-human collaborations about how people exaggerate their motion to disambiguate their goal or intention \cite{pezzulo2013human}.

\emph{Our ability to coordinate with each other relies heavily on predicting each others' intentions \cite{pezzulo2013human}. Modeling human inferences about intentions enables robots to purposefully deviate from efficiency in order to maximally clarify their intentions.}

\prg{Communicating Robot Reward Parameters}
More recently, we've been exploring how $\theta_{\R}$ does not have to be restricted to a goal. Much like how in \sref{sec:inferhuman} we inferred not just human goals, but also more generally human reward function parameters, here too the robot can express not just goals.

\begin{figure}[h]
\includegraphics[width=\columnwidth]{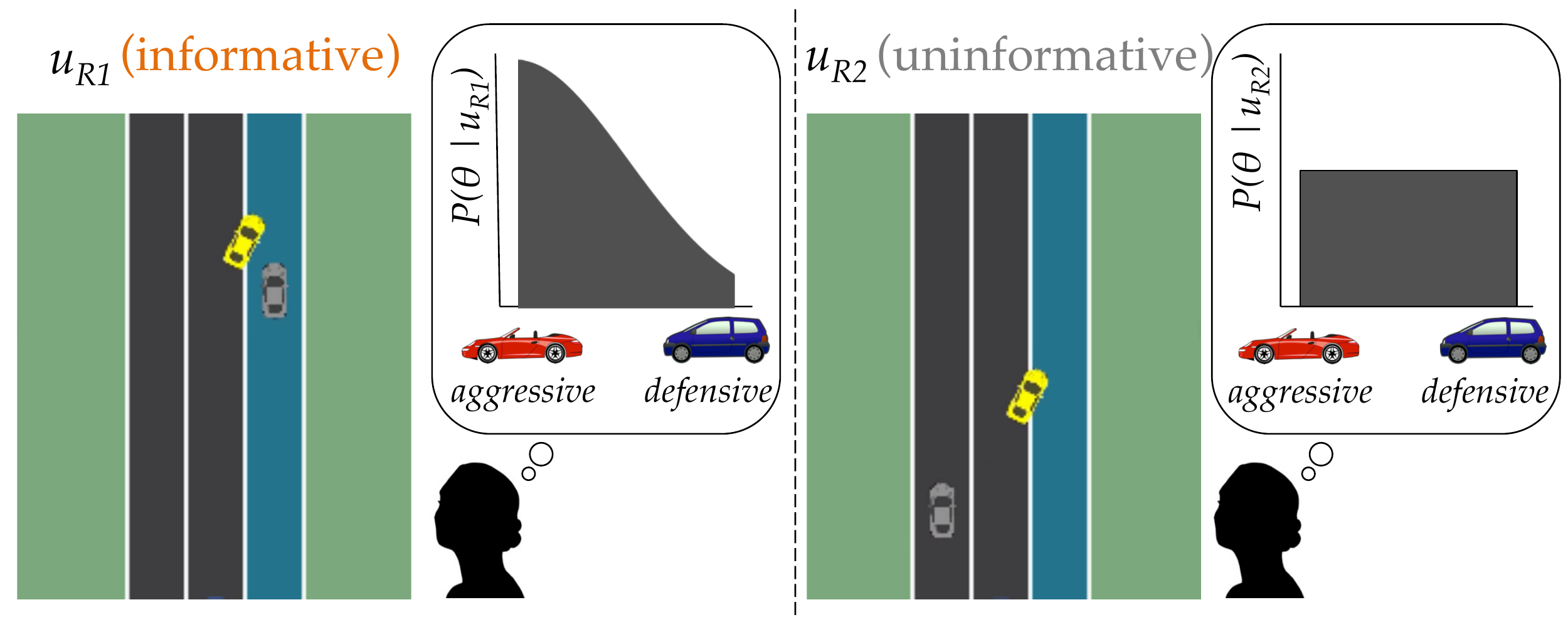}
\caption{The robot models the person as running Bayesian Inference to infer its objective parameters $\theta$. It chooses its actions such that they maximally convey information about $\theta_{\R}$ -- in this case that it prioritize aggressive driving/efficiency over safety.}
\label{fig:rewards} 
\end{figure}

In \cite{Huang_expressrewards}, we studied how an autonomous car can plan behavior that is informative of its reward function. The car decides to show an environment in which the optimal trajectory merges closely in front of another car (\figref{fig:rewards}, left), as opposed to an environment where merging into a lane away from the other car is optimal (\figref{fig:rewards}, right) -- it finds behavior that is informative about the fact that its reward is rather aggressive as a driving style, prioritizing efficiency over safety. 

\emph{As robots get more complex, understanding and verifying their reward functions is going to become more and more important to end-users. Modeling human inferences about reward parameters enables robots to choose actions sequences that are communicative of the true reward parameters.}

\begin{figure}[h]
\includegraphics[width=\columnwidth]{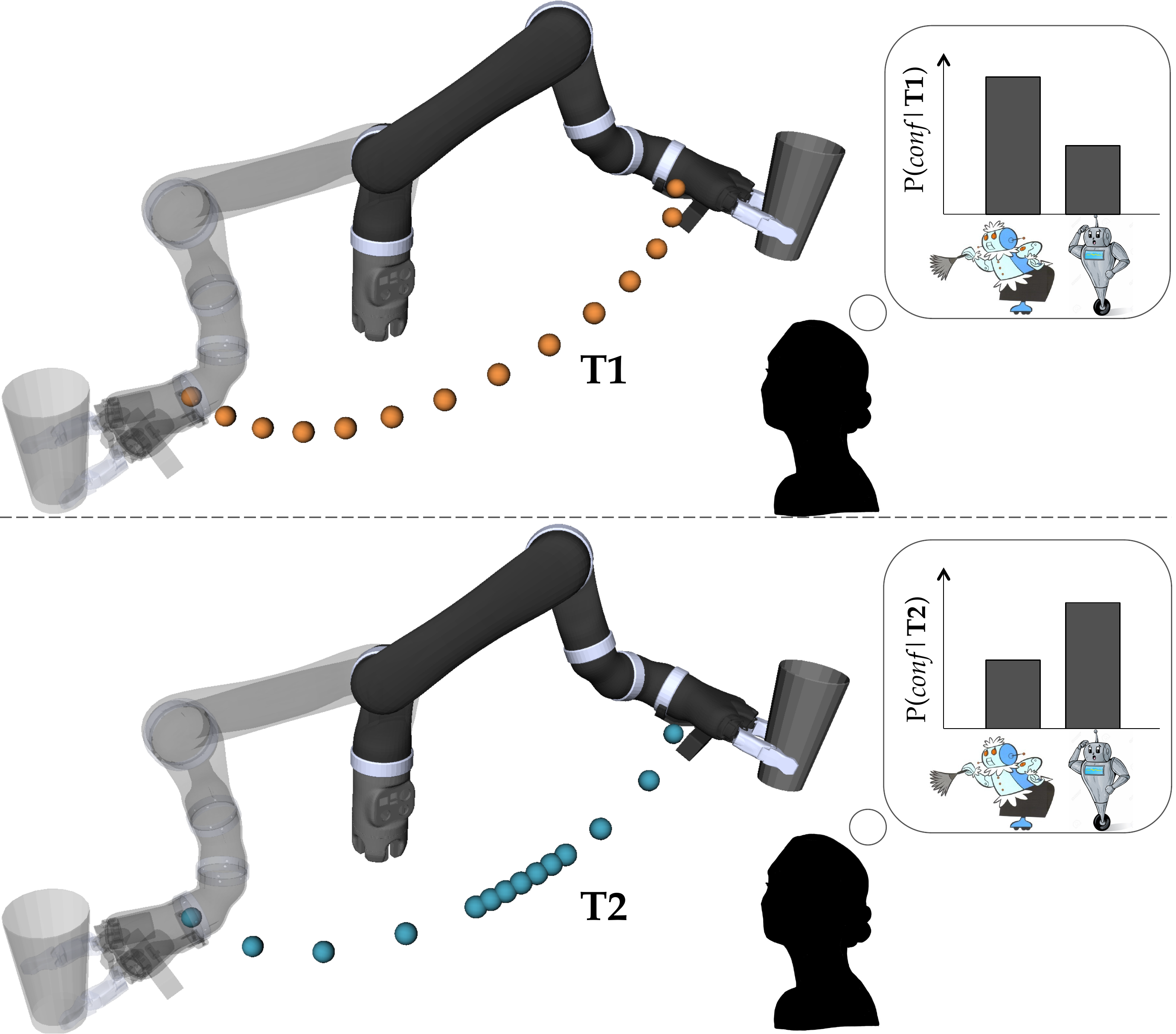}
\caption{Different motion timings communicate different levels of robot confidence.}
\label{fig:timing}
\end{figure}

\prg{Communicating Confidence}
Even more recently, we have been exploring spaces of $\theta$s beyond even reward parameters. Robot actions implicitly communicate about many different aspects of robot internal state. We have found that people observe robot actions and make attributions about its confidence (\figref{fig:timing}), or about the weight of the object that the robot is carrying \cite{Zhang_HRI}. 

\section{Discussion}
This writeup synthesized our findings in integrating mathematical models of human state and action into robot planning of physical behaviors for interactive tasks. We focused on rational models of human behavior in a two-player game, showing how different approximations to the solution lead to different robot behaviors.

A first set of approximations assume that the person has access to what the robot will do, and the robot has to the person's overall reward or utility function. Still, we found that the robot generates behaviors that adapt to the person, that guide the person towards better performance in the task, or that account for the influence the robot will have on what the person ends up doing. We saw robots handing over objects to compensate for people's tendencies to just grasp them in the most comfortable way, and cars being more effective on the road by triggering responses from other drivers.

More sophisticated approximations accounted for the fact that different people have different reward functions, and showed that the robot can actively estimate relevant parameters online, leading to interesting coordination behaviors, like cars deciding on trajectories that look like inching forward at intersections or nudging into lanes to probe whether another driver will let them through.  

Finally, even further approximations acknowledge that people will need to make predictions about the robot, in the same way that the robot makes predictions about people. This leads to robots that are more transparent, communicating their reward function (e.g. their driving style) through the way they act.

This work is limited in many ways, including the fact that as models of people get more complex, it becomes harder to generate robot behavior in real time (especially behavior that escapes poor local optima). However, it is exciting to see the kinds of coordination behaviors that we typically need to hand-craft starting to emerge out of low-level planning \emph{directly in the robot's control space}. This requires breaking outside of the typical AI paradigm, and formally reasoning about people's internal states and behavior.

%
%
%
%
%
%
%
%
%
%
%

{\small
\bibliographystyle{plainnat}
\bibliography{refs}}

\begin{thebibliography}{25}
\providecommand{\natexlab}[1]{#1}
\providecommand{\url}[1]{\texttt{#1}}
\expandafter\ifx\csname urlstyle\endcsname\relax
  \providecommand{\doi}[1]{doi: #1}\else
  \providecommand{\doi}{doi: \begingroup \urlstyle{rm}\Url}\fi

\bibitem[Bajcsy et~al.(2017)Bajcsy, Losey, O'Malley, and
  Dragan]{bajcsy2017phri}
Andrea Bajcsy, Dylan Losey, Martia O'Malley, and Anca Dragan.
\newblock Learning robot objectives from physical human interaction.
\newblock In \emph{in review}, 2017.

\bibitem[Baker et~al.(2007)Baker, Tenenbaum, and Saxe]{baker2007goal}
Chris~L Baker, Joshua~B Tenenbaum, and Rebecca~R Saxe.
\newblock Goal inference as inverse planning.
\newblock In \emph{Proceedings of the Cognitive Science Society}, volume~29,
  2007.

\bibitem[Bestick et~al.(2016)Bestick, Bajcsy, and
  Dragan]{Bestick_ISER_influence}
A.~Bestick, R.~Bajcsy, and A.D. Dragan.
\newblock Implicitly assisting humans to choose good grasps in robot to human
  handovers.
\newblock In \emph{International Symposium on Experimental Robotics (ISER)},
  2016.

\bibitem[Csibra and Gergely(2007)]{csibra2007obsessed}
Gergely Csibra and Gy{\"o}rgy Gergely.
\newblock ?obsessed with goals?: Functions and mechanisms of teleological
  interpretation of actions in humans.
\newblock \emph{Acta psychologica}, 124\penalty0 (1):\penalty0 60--78, 2007.

\bibitem[Dragan and Srinivasa(2012)]{Dragan_RSS_teleop}
A.D. Dragan and S.S. Srinivasa.
\newblock Formalizing assistive teleoperation.
\newblock In \emph{Robotics: Science and Systems (R:SS)}, 2012.

\bibitem[Dragan and Srinivasa(2013)]{Dragan_RSS_legibility}
A.D. Dragan and S.S. Srinivasa.
\newblock Generating legible motion.
\newblock In \emph{Robotics: Science and Systems (R:SS)}, 2013.

\bibitem[Fisac et~al.(2016)Fisac, Liu, Harick, Hedrick, Sastry, Griffiths, and
  Dragan]{Fisac_WAFR_tpred}
J.~Fisac, C.~Liu, J.~Harick, K.~Hedrick, S.~Sastry, T.~Griffiths, and A.D.
  Dragan.
\newblock Generating plans that predict themselves.
\newblock In \emph{Workshop on the Algorithmic Foundations of Robotics (WAFR)},
  2016.

\bibitem[Hadfield-Menell et~al.(2017{\natexlab{a}})Hadfield-Menell, Dragan,
  Abbeell, and Russell]{Menell_offswitch}
D.~Hadfield-Menell, A.D. Dragan, P.~Abbeell, and S.~Russell.
\newblock The off-switch game.
\newblock In \emph{International Joint Confernece on Artificial Intelligence
  (IJCAI)}, 2017{\natexlab{a}}.

\bibitem[Hadfield-Menell et~al.(2016)Hadfield-Menell, Dragan, Abbeel, and
  Russell]{hadfield2016cooperative}
Dylan Hadfield-Menell, Anca Dragan, Pieter Abbeel, and Stuart~J Russell.
\newblock Cooperative inverse reinforcement learning.
\newblock In \emph{Advances in Neural Information Processing Systems}, pages
  3909--3917, 2016.

\bibitem[Hadfield-Menell et~al.(2017{\natexlab{b}})Hadfield-Menell, Russell,
  Abbeel, and Dragan]{hadfield2017ird}
Dylan Hadfield-Menell, Stuart~J Russell, Pieter Abbeel, and Anca Dragan.
\newblock Inverse reward design.
\newblock In \emph{in review}, 2017{\natexlab{b}}.

\bibitem[Hedden and Zhang(2002)]{hedden2002you}
Trey Hedden and Jun Zhang.
\newblock What do you think i think you think?: Strategic reasoning in matrix
  games.
\newblock \emph{Cognition}, 85\penalty0 (1):\penalty0 1--36, 2002.

\bibitem[Huang et~al.(2017)Huang, Abbeel, and Dragan]{Huang_expressrewards}
S.~Huang, P.~Abbeel, and A.D. Dragan.
\newblock Enabling robots to communicate their objectives.
\newblock In \emph{Robotics: Science and Systems (RSS)}, 2017.

\bibitem[Liu et~al.(2016)Liu, Harick, Fisac, Dragan, Hedrick, Sastry, and
  Griffiths]{Chang_AAMAS_goalinference}
C.~Liu, J.~Harick, J.~Fisac, A.D. Dragan, K.~Hedrick, S.~Sastry, and
  T.~Griffiths.
\newblock Goal inference improves objective and perceived performance in
  human-robot collaboration.
\newblock In \emph{Autonomous Agents and Multiagent Systems (AAMAS)}, 2016.

\bibitem[Milli et~al.(2017)Milli, Hadfield-Menell, Dragan, Abbeell, and
  Russell]{Milli_obedience}
S.~Milli, D.~Hadfield-Menell, A.D. Dragan, P.~Abbeell, and S.~Russell.
\newblock Should robots be obedient?
\newblock In \emph{International Joint Conference on Artificial Intelligence
  (IJCAI)}, 2017.

\bibitem[Ng et~al.(2000)Ng, Russell, et~al.]{ng2000algorithms}
Andrew~Y Ng, Stuart~J Russell, et~al.
\newblock Algorithms for inverse reinforcement learning.
\newblock In \emph{Icml}, pages 663--670, 2000.

\bibitem[Pezzulo et~al.(2013)Pezzulo, Donnarumma, and Dindo]{pezzulo2013human}
Giovanni Pezzulo, Francesco Donnarumma, and Haris Dindo.
\newblock Human sensorimotor communication: A theory of signaling in online
  social interactions.
\newblock \emph{PLoS One}, 8\penalty0 (11):\penalty0 e79876, 2013.

\bibitem[Ramachandran and Amir(2007)]{ramachandran2007bayesian}
Deepak Ramachandran and Eyal Amir.
\newblock Bayesian inverse reinforcement learning.
\newblock \emph{Urbana}, 51\penalty0 (61801):\penalty0 1--4, 2007.

\bibitem[Rubinstein(1998)]{rubinstein1998modeling}
Ariel Rubinstein.
\newblock \emph{Modeling bounded rationality}.
\newblock MIT press, 1998.

\bibitem[Sadigh et~al.(2016{\natexlab{a}})Sadigh, Sastry, Seshia, and
  Dragan]{Sadigh_IROS_infogather}
D.~Sadigh, S.~Sastry, S.~Seshia, and A.D. Dragan.
\newblock Information gathering actions over human internal state.
\newblock In \emph{International Conference on Intelligent Robots and Systems
  (IROS)}, 2016{\natexlab{a}}.

\bibitem[Sadigh et~al.(2016{\natexlab{b}})Sadigh, Sastry, Seshia, and
  Dragan]{Sadigh_RSS_driving}
D.~Sadigh, S.~Sastry, S.~Seshia, and A.D. Dragan.
\newblock Planning for autonomous cars that leverages effects on human drivers.
\newblock In \emph{Robotics: Science and Systems (R:SS)}, 2016{\natexlab{b}}.

\bibitem[Sadigh et~al.(2017)Sadigh, Dragan, Sastry, and
  Seshia]{Sadigh_comparison}
D.~Sadigh, A.D. Dragan, S.~Sastry, and S.~Seshia.
\newblock Active preference-based learning of reward functions.
\newblock In \emph{Robotics: Science and Systems (RSS)}, 2017.

\bibitem[Sodian et~al.(2004)Sodian, Schoeppner, and Metz]{sodian2004infants}
Beate Sodian, Barbara Schoeppner, and Ulrike Metz.
\newblock Do infants apply the principle of rational action to human agents?
\newblock \emph{Infant Behavior and Development}, 27\penalty0 (1):\penalty0
  31--41, 2004.

\bibitem[Thomaz et~al.(2016)Thomaz, Hoffman, Cakmak,
  et~al.]{thomaz2016computational}
Andrea Thomaz, Guy Hoffman, Maya Cakmak, et~al.
\newblock Computational human-robot interaction.
\newblock \emph{Foundations and Trends{\textregistered} in Robotics},
  4\penalty0 (2-3):\penalty0 105--223, 2016.

\bibitem[Zhang et~al.(2017)Zhang, Hatfield-Menell, Nagabadi, and
  Dragan]{Zhang_HRI}
A.~Zhang, D.~Hatfield-Menell, A.~Nagabadi, and A.D. Dragan.
\newblock Expressive robot motion timing.
\newblock In \emph{International Conference on Human-Robot Interaction (HRI)},
  2017.

\bibitem[Ziebart et~al.(2008)Ziebart, Maas, Bagnell, and
  Dey]{ziebart2008maximum}
Brian~D Ziebart, Andrew~L Maas, J~Andrew Bagnell, and Anind~K Dey.
\newblock Maximum entropy inverse reinforcement learning.
\newblock In \emph{AAAI}. Chicago, IL, USA, 2008.

\end{thebibliography}

\end{document}